\documentclass{article}
\usepackage{ijcai23}

\usepackage{times}
\usepackage{soul}
\usepackage{url}
\usepackage[hidelinks]{hyperref}
\usepackage[utf8]{inputenc}
\usepackage[small]{caption}
\usepackage{graphicx}
\usepackage{amsmath}
\usepackage{amsthm}
\usepackage{booktabs}
\usepackage{algorithm}
\usepackage{algorithmic}
\usepackage[switch]{lineno}
\usepackage{xcolor}
\usepackage{makecell}
\usepackage{amssymb}
\def \pan#1 {{\color{blue}Pan: #1}}

\title{Optimizing Crop Management with Reinforcement Learning and Imitation Learning}

\author{
Ran Tao$^1$
\and
Pan Zhao$^1$\and
Jing Wu$^{1}$\and
Nicolas F. Martin$^1$\and
Matthew T. Harrison$^2$\and
Carla Ferreira$^3$\and
Zahra Kalantari$^4$\And
Naira Hovakimyan$^1$
\affiliations
$^1$University of Illinois at Urbana-Champaign\\
$^2$University of Tasmania\\
$^3$Stockholm University\\
$^4$ZKTH Royal Institute of Technology
\emails
\{rant3, panzhao2,jingwu6,nfmartin,nhovakim\}@illinois.edu,
matthew.harrison@utas.edu.au,
carla.ferreira@natgeo.su.se,
zahrak@kth.se
}

\begin{document}

\maketitle

\begin{abstract}
Crop management has a significant impact on crop yield, economic profit, and the environment. Although management guidelines exist, finding the optimal management practices is challenging. Previous work used reinforcement learning (RL) and crop simulators to solve the problem, but the trained policies either have limited performance or are not deployable in the real world. In this paper, we present an intelligent crop management system that optimizes nitrogen fertilization and irrigation simultaneously via RL, imitation learning (IL), and crop simulations using the Decision Support System for Agrotechnology Transfer (DSSAT). We first use deep RL, in particular, deep Q-network, to train management policies that require a large number of state variables from the simulator as observations (denoted as full observation). We then invoke IL to train management policies that only need a few state variables that can be easily obtained or measured in the real world (denoted as partial observation) by mimicking the actions of the RL policies trained under full observation. Simulation experiments using the maize crop in Florida (US) and Zaragoza (Spain) demonstrate that the trained policies from both RL and IL techniques achieved more than 45\% improvement in economic profit while causing less environmental impact compared with a baseline method. Most importantly, the IL-trained management policies are directly deployable in the real world as they use readily available information.
\end{abstract}

\section{Introduction}
The agricultural industry worldwide is facing significant challenges regarding food security and nutrition and sustainable agriculture,  which relate to one of the United Nations Sustainable Development Goals (SDGs), Zero Hunger. Humanity must produce food for a population expected to reach 9.6 billion by 2050, and simultaneously reduce environmental impacts, including ecosystem degradation and high greenhouse gas emissions \cite{searchinger2019creating}. There are numerous management factors influencing crop yield and environmental impact, among which nitrogen (N) fertilization and irrigation are two of the most significant ones \cite{reddy2003crop}. Based on empirical experience and existing agricultural studies, local best management practices for N fertilization and irrigation exist among farmers, e.g. \cite{wright2022field,skhiri2012impact}. However, it remains to be seen whether the current management practices are optimal and whether these strategies perform well in the presence of changes in climate and market conditions. Thus, new methods are urgently needed to help farmers build cost-effective and readily deployable systems \cite{ara2021application} that provide optimal management policies given a particular condition (including climate, yield price, management cost, etc.) and a target (e.g., maximum economic profit). 
Since reinforcement learning (RL) has already demonstrated extraordinary ability in solving tasks involving sequential decision making (SDM), e.g., in robotics and games \cite{kaufmann2018droneracing-rl,mnih2015human-deepRL}, we see the great potential of RL in optimizing crop management, which is essentially an SDM problem. As numerous interactions between the RL agent and the environment are required for policy training, it is impossible to implement field trial-based methods. Therefore, using crop models to simulate the crop and environment and interact with the RL agent \cite{palmer2013influence,attonaty1997using} seems the only realistic solution. 

Recently, the authors of \cite{wu2022optimizing} proposed to train management policies for N management using deep RL, the Decision Support System for Agrotechnology Transfer (DSSAT) \cite{jones2003dssat}, one of the most widely used crop models in the world, and Gym-DSSAT \cite{romain2022gym}. Their trained policies under full observations outperformed a baseline policy by achieving a higher yield or a similar yield with less N fertilizer input. 
However, there are a few limitations to this study. First, only N fertilization management was optimized. Also, only one reward function was adopted in the training and it is unclear whether their framework works well for various situations with different goals. Most importantly, their trained policies under full observation are not implementable in the real world as they need information that is not accessible by farmers, such as nitrate leaching and plant N uptake. Although the authors of \cite{wu2022optimizing} conducted experiments on policy training under partial observation, using only easily obtained or measurable state variables in reality, the training results could not outperform the baseline policy, let alone the ones trained under full observation. These limitations motivate the present paper.
\begin{figure}[t]
  \centering
  \includegraphics[width=1\linewidth]{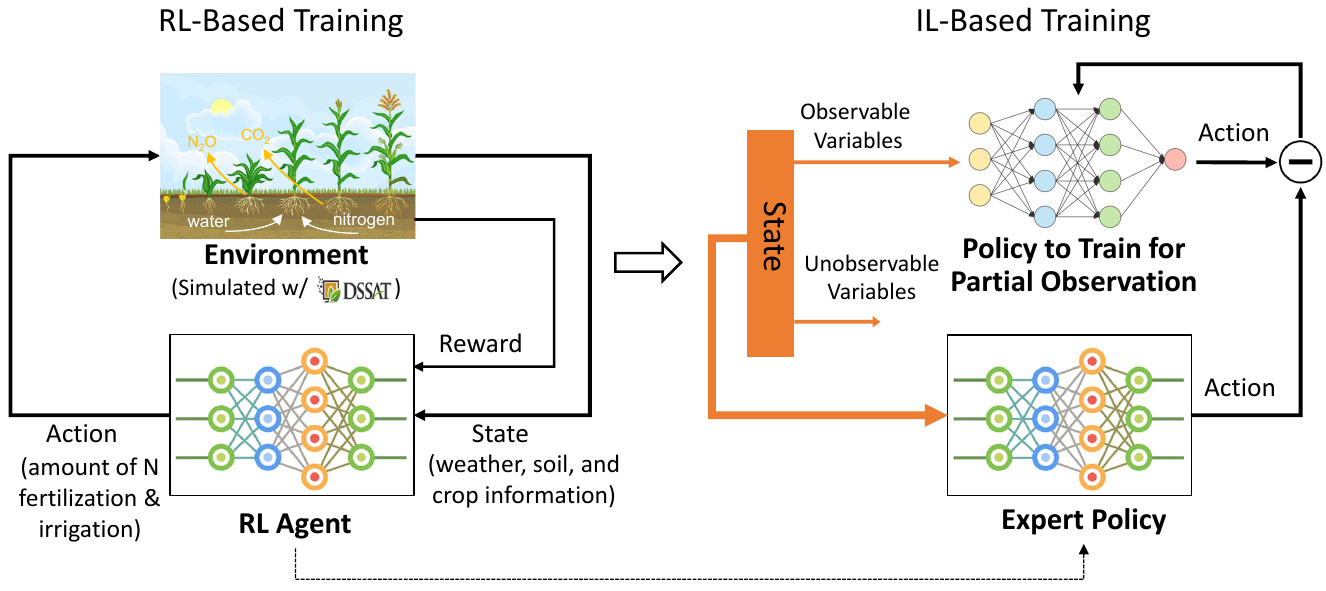}
   \caption{Framework of the intelligent crop management system using RL and IL. We first use RL to train management policies under full observation. Then, the RL-trained policies are used as the expert to train policies under partial observation with IL.}
   \label{fig:Framework}
\end{figure}

In this paper, we present an intelligent crop management framework, depicted in Figure \ref{fig:Framework}, that generates {\it deployable} and {\it adaptable} management policies based on RL, imitation learning (IL), crop simulations via DSSAT, and Gym-DSSAT. Our proposed framework combines RL on crop management, from \cite{wu2022optimizing}, and IL on solving the problem of partial observation given an expert policy. We validate the performance of the proposed framework using the maize crop in Florida, US, and Zaragoza, Spain. For both cases, the trained policies outperform a baseline management policy following either a maize production guide recommended by domain experts or survey data on real-world management practices from maize farmers. Compared with \cite{wu2022optimizing}, this paper advances the state of the art in the following directions. First, unlike existing works considering either N management \cite{wu2022optimizing} or irrigation management \cite{sun2017rl-irrigation}, we study joint N and irrigation management.
Second, we investigate the RL-based policy training with different reward functions that represent different tradeoffs among crop yield, resource usage, and environmental impact in terms of nitrate leaching during the crop growth cycle. Most importantly, we leverage IL as a new tool to find the optimal management policies that require only state variables that can be easily obtained or measured in the real world. 
In addition, the efficacy of the IL-trained policies is evaluated in the presence of measurement noises to mimic the real-world situation when farmers obtain state variables from either weather forecasts or measurements. 
As a result, the path to the deployment of our intelligent crop management system is paved. 

\section{Related Work}
\subsection{Crop Management with RL}
By viewing crop management as an SDM problem, to be more specific, as a Markov decision process (MDP) problem, researchers have applied RL to find the optimal crop management policies from simulators; however, this direction is very much in its infancy. A simple RL method optimizing wheat management in France was presented by \cite{garcia1999rl-wheat}. The authors of \cite{sun2017rl-irrigation} researched the optimization of the irrigation management for the maize crop in Texas, US. The previous two papers did not include a comprehensive study due to their small state spaces and action spaces. For instance, only one state variable was included for RL training in \cite{sun2017rl-irrigation}. The authors of \cite{ashcraft2021ml-aided} utilized the proximal policy optimization (PPO) algorithm to optimize management policy for fertilizer and irrigation; however, the results are not impressive as the trained policies cannot outperform some baseline methods in the simulation. Recently, a more profound study on applying RL to train crop management policies was conducted by \cite{wu2022optimizing}, which focused on N fertilization for maize in Florida and Iowa. The authors included a larger state space, and the trained policies under full observation achieved better outcomes than a baseline method. A more comprehensive review of research work in RL-based crop management can be found in \cite{gautron2022reinforcement}.

\subsection{Crop Models for RL}
\label{section:dssat}
As the real-world farm experiments are laborious, time-consuming, and cost-inefficient, crop models are becoming essential tools for estimating the crop responses to climate change and management practices on crop production \cite{zhao2019simple}. Until now, multiple crop simulation models have been developed for major crops \cite{bassu2014various,asseng2013uncertainty}. According to \cite{jones2017brief}, APSIM and DSSAT are the two most widely used crop models. In addition to providing accurate crop production estimates in relation to climate, genotype, soil, and management factors, these two crop models are open-source for the agricultural community and being updated constantly. 

For currently available crop models, the users need to pre-define the management practice before the start of a simulation. However, RL aims to find optimal policies that decide the management practices in (near) real-time according to the current weather, plant, and soil conditions. Given this concern, efforts have been made to enable real-time communication between an RL agent and the crop environment during a simulation. The authors of \cite{overweg2021cropgym} developed the CropGym environment for training fertilization policies following the convention of the Open AI Gym \cite{brockman2016openai}, an environment widely used for developing and testing RL agents. Another interface based on the SIMPLE crop model for Open AI Gym is proposed in \cite{zhao2019simple} for russet potatoes. While these works prove the feasibility of deploying RL methods for crop management, the simulation models used are over-simplified without detailed information on the crop and environment. In comparison, Gym-DSSAT, based on the widely used DSSAT, is developed for training management policies with RL \cite{romain2022gym}. It enables daily interactions between an RL agent and the simulated crop environment in DSSAT and has been leveraged to optimize N management \cite{wu2022optimizing}.

\subsection{Imitation Learning for Policy Training}
IL seeks to learn a policy by imitating the behavior of an expert and it has been widely applied in the field of robotics especially for autonomous vehicles. The authors of \cite{zhang2016query} applied IL to learn autonomous driving policies using an expert driver as the learning target. The authors of \cite{zhang2016learning} alternatively used the model predictive controller (MPC) as the expert to train a control policy with limited observation. Both of these works indicate the promising ability of IL in learning good policies given an expert. \cite{zhang2016learning} further demonstrates the ability of IL in learning good policies under partial observation, which is usually difficult to achieve with standard RL due to the less available information compared with the full observation case. However, to the best of our knowledge, there has been no reported work on applying IL for crop management.

\section{Methods}

\subsection{MDP Problem Formulation for Crop Management}
Nitrogen fertilization and irrigation management is formulated as a finite MDP here. On each day $t$, the agent receives the state of the environment, $s_t$, and chooses the action $a_t$ from the action space $\mathcal{A}$ based on some policy $\pi(s_t,\theta_t)$, where $\theta_t$ is the parameter of the policy at current day. $s_t$ contains information related to the weather, plant, and soil at given day from the simulator. $a_t$ consists of the amount of N fertilizer input, $N_t$, and the amount of irrigation water, $W_t$, to be applied on that day. Given $s_t$ and $a_t$, the reward $r_t (s_t,a_t)$ is defined as:
\begin{equation}\label{eq:reward}
    r_t(s_t,a_t) \!=\! \left\{\hspace{-2mm}
    \begin{array}{ll}
         w_1Y\!-\! w_2 N_t \!-\! w_3 W_t \! - \! w_4 N_{l,t} \! & \hspace{-2mm}\textup{if }  \textup{harvest at $t$,}    \\
        \!- w_2 N_t \!-\! w_3 W_t \! - \! w_4 N_{l,t}  & \hspace{-2mm} \textup{otherwise,} 
    \end{array}
    \right.
\end{equation}
where $w_1,w_2,w_3,w_4,Y,N_{l,t}$ are four custom weight factors, yield at harvest, and the amount of nitrate leaching at given day respectively. $Y$ and $N_{l,t}$ are variables from the state $s_t$. Given the state $s_t$ and a design of the reward function represented by $w_1,w_2,w_3,w_4$, the goal of the agent is to find the optimal policy $\pi(s_t,\theta_t)$ which renders $a_t$ and maximizes the future discounted return, which is defined as $R_t = \sum_{\tau=t}^T\gamma^{\tau-t}r_\tau$ representing the sum the reward at current day from applying $a_t$, and discounted future rewards with factor $\gamma$ following this policy.

\subsection{Policy Training using RL}
Deep Q-network (DQN), a deep RL algorithm, is selected for agent training to learn the optimal policy that maximizes the future discounted return $R_t$. With DQN, a deep neural network is used to represent the action-value function, also known as the Q function \cite{mnih2015human-deepRL}, and thus we call the network as a Q-network. The Q function of a policy $\pi$ is defined as $Q^\pi(s,a) = \mathop{\mathbb{E}}[R_t|s_t=s,a_t=a,\pi]$, which measures the expected future discounted return obtained from state $s$ by taking action of $a$ and following policy $\pi$ afterwards. We keep updating the parameters of the Q-network to find the optimal Q function, $Q^\star(s,a)$, which represents the optimal return that can be gained from state $s$ by taking action $a$ and following the optimal policy afterwards. Given $Q^\star$ and $s_t$, a greedy policy defined as $a_t^\star = \max_{a\in \mathcal{A}}Q^\star(s_t,a)$ is used by the agent to determine the optimal action. Since the Q function determines the policy of the agent, training of the Q-network is same as the training of the policy. We update the parameter of the Q-network at iteration $i$, denoted by $\theta_i$, by minimizing the loss function: 
\begin{equation}
L_i(\theta_i) \! \triangleq
  \!\! \mathop{\!\mathbb{E}}_{(s,a,r,s')}\!\left[r\!+\!\gamma \max_{a'\in \mathcal{A}}Q(s'\!,a';\theta_i^{-}) \!-\! Q(s,a;\theta_i)\!\right]\!,
\end{equation}
where $s,a,r,s'$ and $\gamma$ denote state, action, reward of $s$ and $a$, next state, and discount factor respectively, and $\theta_i^{-}$ denotes the parameters of a previously defined target network. The tuples $(s,a,r,s')$ are randomly chosen from the replay buffer, which is a memory base of previously generated tuples $(s,a,r,s')$ during training. The RL-trained policies under full observation are later used as the expert in IL. 

\subsection{Crop Simulations with Gym-DSSAT}
Similar to \cite{wu2022optimizing}, we leverage Gym-DSSAT \cite{romain2022gym}, a Gym interface for DSSAT that enables the agent to interact with the simulated environment (i.e., reading the weather, soil, and crop information and applying management practices) on a daily basis. For more details about DSSAT and Gym-DSSAT, readers can refer to Section \ref{section:dssat}.

\subsection{Policy Training using IL}
Imitation learning trains the agent to perform a task by mimicking the behavior of an expert \cite{osa2018algorithmic}. As opposed to learning from scratch by trial-and-error in RL \cite{hussein2017imitation}, with IL the agent learns a mapping between the observations and desired actions determined by the expert, which simplifies the learning process of complex problems. For the crop management problem, not all state variables from the simulator can be achieved or measured by farmers. Thus, the agent should only utilize state variables that are accessible to farmers during real-world deployment. Given any state $s$, denote $s^o$ as the observable state which contains variables from $s$ that are observable or measurable by farmers. Under partial observation, on each day $t$, the agent receives $s^o_t$. The goal of the agent is to learn an optimal policy $\pi(s^o_t,\theta)$ that outputs an action $a^o_t$ identical to $a_t$, where $a_t$ is the action determined by the expert given $s_t$. Behavior cloning, the simplest form of IL, can be applied to train the policy under partial observation as follows. We first collect demonstrations, state-action pairs $(s, a)$, from the expert policy and store them into a dataset $\mathcal{D}$. Then, the policy network of the agent is updated by minimizing the loss function
\begin{align}
\label{eq:IL}
    L(\theta) = \sum_{(s,a)\in \mathcal{D}}\|\pi(s^o,\theta)-a\|.
\end{align}
The loss function represents the difference between the output of the policy network with $s^o$ as the input, and the action $a$ determined by the expert policy given $s$.
\begin{table*}[t]
    \centering
    \footnotesize
    \caption{State variable description}
    \label{table:state-description}
    \begin{tabular}{l|l|l}\toprule
    Variable      & Description                                                                       & Included in Partial Observation Study?\\ \midrule
    cumsumfert & cumulative nitrogen fertilizer applications (kg/ha)                               & Yes     \\ \hline
    dap        & days after simulation started                                                     & Yes     \\ \hline
    dtt        & growing degree days for current day (C/d)                                         & Yes      \\ \hline
    istage     & DSSAT maize growing stage                                                         & Yes       \\ \hline
    vstage     & vegetative growth stage (number of leaves)                                        & Yes       \\ \hline
    pltpop     & plant population density (plant/m2)                                               & Yes      \\ \hline
    rain       & rainfalls for the current day (mm/d)                                              & Yes       \\ \hline
    srad       & solar radiations during the current day (MJ/m2/d)                                 & Yes       \\ \hline
    tmax       & maximum temparature for current day (C)                                           & Yes       \\ \hline
    tmin       & minimum temparature for current day (C)                                           & Yes       \\ \hline
    sw         & volumetric soil water content in soil layers (cm3 {[}water{]} /   cm3 {[}soil{]}) &  Yes      \\ \hline
    xlai       & plant population leaf area index (m2\_leaf/m2\_soil)                              &  Yes     \\ \hline
    nstres     & index of plant nitrogen stress (unitless)                                         &  No   \\ \hline
    pcngrn     & massic fraction of nitrogen in grains (unitless)                                  &  No     \\ \hline
    swfac      & index of plant water stress (unitless)                                            &  No     \\ \hline
    tleachd    & daily nitrate leaching (kg/ha)                                                    &  No     \\ \hline
    grnwt      & grain weight dry matter (kg/ha)                                                   &  No     \\ \hline
    cleach     & cumulative nitrate leaching (kg/ha)                                               &  No     \\ \hline
    cnox       & cumulative nitrogen denitrification (kg/ha)                                       &  No     \\ \hline
    tnoxd      & daily nitrogen denitrification (kg/ha)                                            &  No     \\ \hline
    trnu       & daily nitrogen plant population uptake (kg/ha)                                    &  No     \\ \hline
    wtnup      & cumulative plant population nitrogen uptake (kg/ha)                               &  No     \\ \hline
    topwt      & top weight (kg/ha)                                       &   No    \\ \hline
    es         & actual soil evaporation rate (mm/d)                                               &  No     \\ \hline
    runoff     & calculated runoff (mm/d)                                                          &  No     \\ \hline
    wtdep      & depth to water table (cm)                                                         &  No     \\ \hline
    rtdep      & root depth (cm)                                                                   &  No     \\ \hline
    totaml     & cumulative ammonia volatilization (kgN/ha)                                        &  No     \\ \bottomrule
    
\end{tabular}
\end{table*}
\section{Experiments and Results}
Experiments on the training of N and irrigation management policies for the maize crop under both full observation and partial observation were conducted with two case studies. The first case study uses the simulated environment of Florida, US in 1982, and the second case study uses the simulated environment of Zaragoza, Spain in 1995.
These two case studies aim to validate the feasibility and benefits of the proposed framework, not for immediate deployment. For real-world deployment of our framework, more details can be found in Section \ref{sec:deployment}.

or each case study, we used DQN to train the RL agent under full observation, and the trained policies were then used as the experts to train management policies under partial observation using IL. We tested the performance of all the trained policies in simulation and compared them with baseline policies. The baseline policy for the Florida case study follows a maize production guide for farmers in Florida \cite{wright2022field} written by domain experts, and the baseline policy for the Zaragoza case comes from the survey data on the real-world management practices of maize farmers in Zaragoza \cite{malik2019dssat,skhiri2012impact}. All experiments included in this paper were conducted at least ten times and the detailed description of state variables can be found in Table \ref{table:state-description}.

Regarding the two baseline methods, the details are listed below.
Florida case:  For N fertilization, 40 kg/ha of fertilizer is applied at the planting date, 40 kg/ha is applied when the maize reaches about 12 inches tall, 150 kg/ha is applied when the maize reaches 15 inches tall, and finally 130kg/ha is applied 4 weeks after. For irrigation, one inch of water is fed every 10 days until the maize reaches 15 inches high, an inch is applied every 7 days before tassel emergence, and finally an inch of water is applied every 3 days until maize maturity.
Spain case: 50 kg N/ha of fertilizer is applied at pre-planting, 100 kg/ha of fertilizer is applied in the first sidedress (15th June), and another 100 kg N/ha is applied in the second sidedress (10th July). Irrigate once every 3 days with 16 mm every time.

\subsection{Policy Training under Full Observation via RL}
\label{sec:RL_full}
DQN was implemented for training the RL agent under full observation. We tested with five different reward functions to demonstrate the adaptability of our framework to different tradeoffs among crop yield, N fertilizer use, irrigation water use, and environmental impact. 

\subsubsection{Implementation Details}
\label{sec:RL FULL}
The neural network of the Q function was designed to have 3 hidden layers with 256 units in each layer. The discrete action space was set as: 
\begin{align}
\label{eq:action space}
    \mathcal A =\{40k  \frac{\textrm{kg}}{\textrm{ha}} \text{ N fertilizer }\; \&\; 6k \frac{\textrm{L}}{m^2} \textrm{ Irrigation water} \}
\end{align}
where $k=0,1,2,3,4$, with a size of 25. This design of the action space includes standard amounts of N fertilizer and irrigation water that farmers can potentially apply in a single day and also provides enough options for finding good policies. The discount factor was set to be 0.99. For updating the neural network, we utilized Pytorch and Adam \cite{kingma2014adam} optimizer with an initial learning rate of 1e-5 and a batch size of 640. 

For each case study, five different functions for $r_t$ from \eqref{eq:reward} were used to train the RL agent and for each reward function, we picked one trained policy for evaluation. The parameters used in each reward function (RF) are listed in Table \ref{table:weight}. 
RF 1 represents the economic profit (\$/ha) that farmers gain based on the approximate price of maize and cost of N fertilizer and irrigation water from \cite{mandrini2022exploring} and \cite{wright2022field}. RFs 2-4 indicate the economic profit under different situations. To be more specific, RF 2 represents the extreme case when irrigation water is free, RF 3 denotes the extreme case when N fertilizer is free, and RF 4 simulates the situation when the price of N fertilizer is doubled. Compared with RFs 1-4 which consider economic profit only, RF 5 includes the additional term of nitrate leaching, an environmental factor. Nitrate leaching is unfavorable since it's the major cause of environmental problems including eutrophication of watercourses and soil degradation \cite{di2002nitrate}. RF 5 is designed with similar weights on yield, N fertilizer usage and irrigation usage, and a much larger weight on nitrate leaching to promote minimal nitrate leaching while obtaining a good economic profit. 

\begin{table}[t]
\caption{Weights used in each reward function (RF) defined by \eqref{eq:reward}}
\label{table:weight}
\small
\centering
\begin{tabular}{l|r|r|r|r|l }\toprule
& \makecell{$w_1$\\ ($Y$)}  & \makecell{$w_2$\\ ($N_t$)} & \makecell{$w_3$ \\($W_t$)} & \makecell{$w_4$ \\($N_{l,t}$)}&Note \\ \midrule
RF 1 & 0.158 & 0.79    & 1.1     & 0 & Economic profit    \\ 
RF 2 & 0.158 & 0.79    & 0       & 0 &  Free water \\ 
RF 3 & 0.158 & 0    & 1.1    & 0 &  Free N fertilizer \\ 
RF 4 & 0.158 & 1.58    & 1.1     & 0  & Double N price  \\ 
RF 5 & 0.2   & 1       & 1       & 5  &  With N Leaching  \\ \bottomrule
\end{tabular}
\end{table}

\begin{table*}[h]
\caption{Evaluation results of the trained policies under full observation and the baseline policy for both case studies. $N_l$ represents the nitrate leaching amount. Trained Policy x indicates the policy trained using reward function (RF) x, and the values in the column of 'RF x' represent the cumulative rewards of each trained policy calculated with RF x. Details of each reward function can be found in Table \ref{table:weight}. For each reward function design, the largest cumulative reward value is shown in bold.}
\centering
\begin{tabular}{l|r|r|r|r|r|r|r|r|r}
\toprule
Florida Case & \makecell{N Input\\ (kg/ha)} & \makecell{Irrigation\\ (L/m$^2$)} & \makecell{$N_l$ \\(kg/ha)} & \makecell{Yield\\ (kg/ha)} &RF 1 &RF 2 &RF 3 &RF 4 &RF 5 \\ \midrule
Baseline Policy        & 360     & 394       & 213      & 10772  &984                         &1417                        &1269                        & 700                         & 338
\\ 
Trained Policy 1       & 200     & 120        & 36      & 10852  &\textbf{1425}                        &1557                       &1538                       &1267                        & \textbf{1673}
\\ 
Trained Policy 2       & 200     & 732         & 59       & 11244    &813                         & \textbf{1619}               &971                        &655                         &1020                    \\ 
Trained Policy 3       & 19920     & 108         & 6205       & 10865    &-1.4e4                       &-1.4e4                     & \textbf{1598}                      &-3.0e4                       &-4.9e4                     \\
Trained Policy 4       & 160     & 102         & 35       & 10358  &1398                        &1510                        &1524                        &\textbf{1272}                        &1635     \\
Trained Policy 5      & 200     & 138         & 39       & 10926 &1417              &1568                        &1575              &1259               &1651     \\ \bottomrule
\end{tabular}
\begin{tabular}{l|r|r|r|r|r|r|r|r|r}
\toprule
Spain Case & \makecell{N Input\\ (kg/ha)} & \makecell{Irrigation\\ (L/m$^2$)} & \makecell{$N_l$ \\(kg/ha)} & \makecell{Yield\\ (kg/ha)} &RF 1 &RF 2 &RF 3 &RF 4 &RF 5 \\ \midrule
Baseline Policy        & 250     & 752       & 4      & 10990  &712 & 1539 & 909 & 514 & 1176
\\ 
Trained Policy 1       & 240     & 330        & 913     & 10477  &\textbf{1103} &1466 & 1292 & 913 & 1525
\\ 
Trained Policy 2       & 200     & 1068         & 5.6       & 10923    &393 & \textbf{1568} & 551 & 235 & 888                    \\ 
Trained Policy 3       & 10640     & 324         & 0       & 10626    &-7083 & -6727 & \textbf{1323} & -1.5e4 & -8839                     \\
Trained Policy 4       & 120     & 336        & 0       & 9601 & 1053 & 1422 & 1147 & \textbf{958} & 1464   \\
Trained Policy 5       & 200     & 390        & 0       & 10589 &1086              &1515                        &1244              &928               &\textbf{1528}      \\ \bottomrule
\end{tabular}
\end{table*}

\begin{figure}[h]
  \centering
  \includegraphics[width=0.9\linewidth]{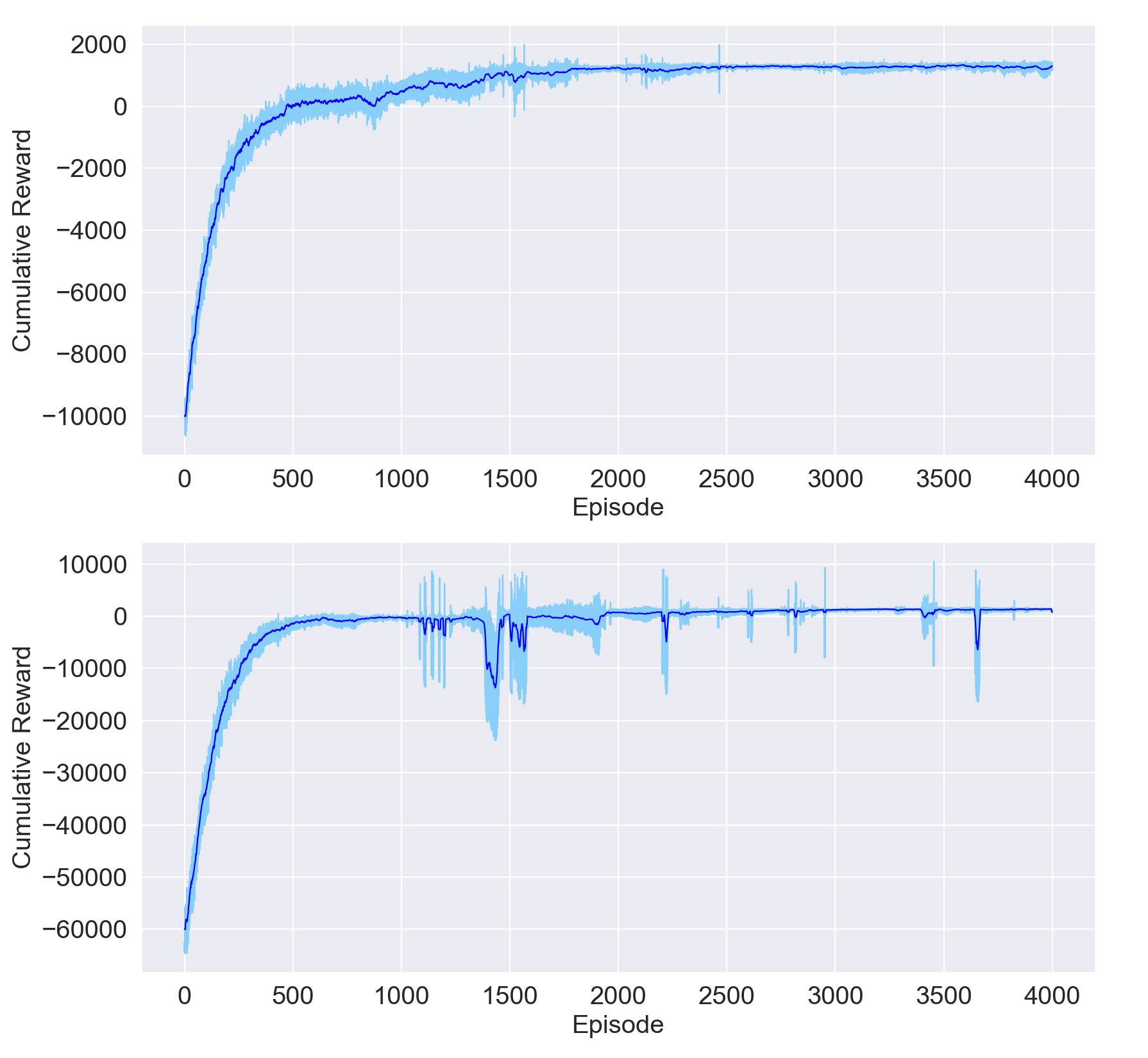}
   \caption{Cumulative reward versus episodes for policy training under RF~1 (Top) and RF~5 (Bottom). Results are averaged over ten trials, with light-blue shaded areas denoting the variance.}
   \label{fig:training_curve}
\end{figure}
\subsubsection{Policy Evaluation}
For illustration purposes, the training curves of the Florida case using RF 1, which only considers economic profit, and RF 5, which includes both economic and environmental considerations, are shown in Figure \ref{fig:training_curve}.
The evaluation results of the trained policies under full observation for both Florida and Zaragoza cases are shown in Table \ref{table:TP_results}. It is worth mentioning that due to the random initialization of the Q-network and the fact that the Q-network gets updated every episode, the policies evaluated may not represent the best one from training. However, the trained policies we picked are representative enough to demonstrate the ability of RL to optimize crop management and the influence of the reward functions on the training results. According to Table \ref{table:TP_results}, the reward function affects the strategy of the trained policy significantly for both case studies. Trained with RF 2 indicating the zero cost of irrigation water, Trained Policy 2 applies the largest amount of irrigation water while keeping the N input low, leading to the highest yield and the largest cumulative reward according to RF 2. Similarly, with a zero cost of N fertilizer in RF 3, Trained Policy 3 from both case studies applies a large amount of N fertilizer with a small amount of irrigation. With RF~4, indicating a doubled price for N fertilizer compared to RF~1, Trained Policy 4 results in a smaller amount of N input compared to Trained Policy 1. Most importantly, given an RF to compute the cumulative rewards of different trained policies, the largest reward is always achieved by the policy trained with this particular RF for both case studies (e.g., Trained policy 1 achieves the highest cumulative reward with RF 1), except the Florida case under RF 5, where Trained policy 5 achieves a cumulative reward slightly smaller than the largest one from Trained Policy 1 but still much larger than the one from the baseline policy. Compared with the baseline policies, the RL-trained trained policies achieve a 45\% and a 55\% increase in terms of profit (RF 1), and almost a 400\% and a 30\% increase in terms of RF 5 for the Florida case and Zaragoza case respectively. The enormous negative values of the cumulative rewards of Trained Policy 3 from both case studies are the results of the large amounts of N input, which are not punished during training with RF 3. In general, the results above demonstrate the ability of RL to optimize crop management under different criteria, at different locations with various environments. Given a specific target represented by a design of the reward function, we can always apply the RL-based training to find the optimal management policies.

\subsubsection{Policy Training of Either N Fertilization or Irrigation}
In the preceding studies, we optimized N fertilization and irrigation simultaneously and the results were compared with baseline policies recommended by domain experts. To examine the benefits of joint optimization of the N and irrigation management, we include additional experiments in this section where we trained management policies for only N fertilization or irrigation with the other practice following the baseline methods, e.g., training an N management policy while following the baseline policy for irrigation management. 
Experiments were conducted using RF 1 and RF 5 for the Florida case study. The results are shown in Table \ref{table:baseline}, which clearly demonstrates the benefits of joint optimization of N fertilization and irrigation management.

\begin{table}[t]
\caption{Performance comparison of the trained policies on both N fertilization and irrigation with the trained policies on either N fertilization or irrigation. The largest values are shown in bold.}
\label{table:baseline}
\small
\centering
\begin{tabular}{l|r|r}\toprule
Florida Case& RF 1 & RF 5 \\ \midrule
Baseline N Fertilization + Baseline Irrigation        & 984    & 338                   \\ 
Training N Fertilization + Training Irrigation      &\textbf{1425}     & \textbf{1673}                    \\
Baseline N Fertilization + Training Irrigation     &1351   & 849                     \\ 
Training N Fertilization + Baseline Irrigation       &1062     &  1217               \\ \bottomrule
\end{tabular}
\end{table}

\subsection{Policy Training under Partial Observation via Imitation Learning}
For RL-based policy training in Section~\ref{sec:RL_full}, we included a large number of state variables from DSSAT as the input to the RL agent. However, most of the state variables used, including nitrate leaching and daily N denitrification, cannot be readily obtained or even measured by farmers with accessible instruments in reality. As a result, there is no way to implement those trained policies on a real farm. A policy that only utilizes state variables that can be easily obtained or measured (partial observation) is necessary for real-world deployment. In this section, we present a method based on IL to train management policies that only use readily available information for decision-making, in which the RL-trained policies under full observation in Section \ref{sec:RL_full} are used as experts to be mimicked. All the state variables needed by the IL-trained policies, can be easily accessed or measured by farmers in the real world, like the weather information, soil water content, which can be measured by a soil moisture meter, and maize growing stage, which can be easily identified by an experienced farmer.

\subsubsection{Limited Performance of RL under Partial Observation}
\begin{figure}[t]
  \centering
  \includegraphics[width=0.9\linewidth]{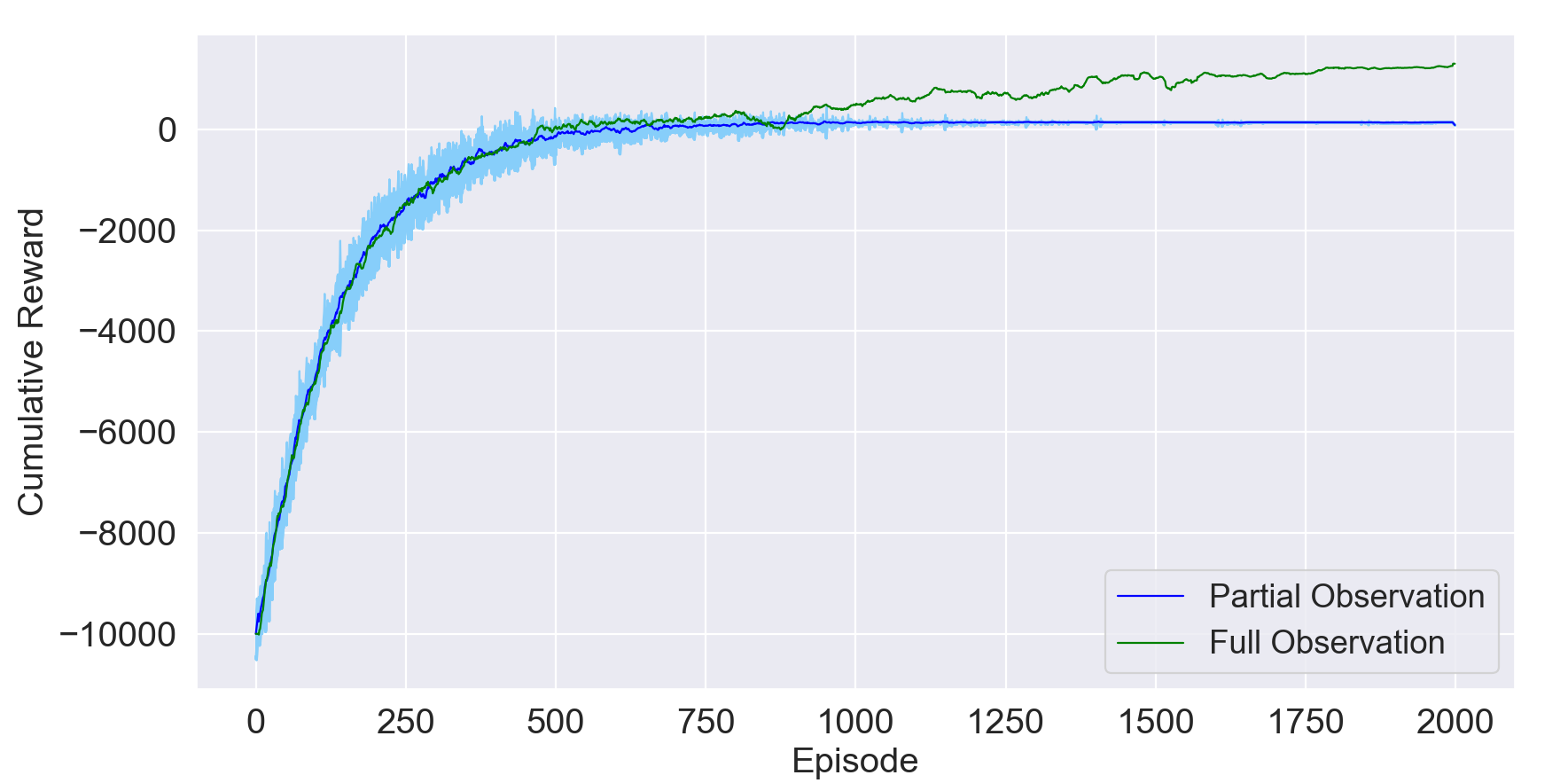}
   \caption{Cumulative reward versus episodes for RL-based policy training under partial observation with RF 1. Results are averaged over ten trials, with the light-blue shaded area denoting the variance.}
   \label{fig:Training_curve_partial}
\end{figure}

\begin{table*}[t]
\caption{Performance comparison between the RL-trained policies (experts) and their corresponding IL-trained policies. $N_l$ represents the N leaching amount. Policies under partial observation can be trained with IL and outperform the baseline policy.}
\centering
\begin{tabular}{l|r|r|r|r|r|r}\toprule
Florida Case& \makecell{N Input \\ (kg/ha)} & \makecell{Irrigation\\ (L/m$^2$)} & \makecell{$N_l$ \\(kg/ha)} &\makecell{Yield\\ (kg/ha)}  &  RF 1   & RF 5   \\ \midrule
Baseline Policy           & 360     & 394       & 212.6      & 10771.5 & 984.4  & 337.6  \\ \hline
RL-Trained Policy 1 (Full)       & 200     & 120         & 35.5       & 10852.4   & 1424.7 & N/A    \\ \hline
IL-Trained Policy 1 (Partial)      & 200     & 138         & 37.0       & 10870.0 & 1407.7& N/A    \\ \hline
RL-Trained Policy 5 (Full)      & 200     & 138         & 39.2       & 10926.1 & N/A    & 1651.0   \\ \hline
IL-Trained Policy 5 (Partial)        & 200     & 138         & 39.2       & 10926.1 & N/A    &  1651.0 \\ \bottomrule
\end{tabular}
\begin{tabular}{l|r|r|r|r|r|r}\toprule
Zaragoza Case& \makecell{N Input \\ (kg/ha)} & \makecell{Irrigation\\ (L/m$^2$)} & \makecell{$N_l$ \\(kg/ha)} &\makecell{Yield\\ (kg/ha)}  &  RF 1   & RF 5   \\ \midrule
Baseline Policy           & 250     & 752      & 4.1      & 10989.9 & 711.7  & 1175.5  \\ \hline
RL-Trained Policy 1 (Full)       & 240     & 330         & 0       & 10476.6   & 1102.7 & N/A    \\ \hline
IL-Trained Policy 1 (Partial)      & 240     & 336         & 0       & 10646.7 & 1123.0& N/A    \\ \hline
RL-Trained Policy 5 (Full)      & 200     & 390         & 0       & 10588.6 & N/A    & 1527.7   \\ \hline
IL-Trained Policy 5 (Partial)        & 200     & 414         & 0       & 10730.5 & N/A    &  1532.1 \\ \bottomrule
\end{tabular}
\label{table:super_partial}
\end{table*}
Under partial observation, we first experimented with RL-based policy training, using the same setup as used in the full-observation case (Section~\ref{sec:RL_full}), except that the state space used here was smaller. Although we tried different hyperparameters, including dimensions of the neural network, learning rates, and batch sizes, all trained policies converged to a single one that applies zero N fertilizer and zero irrigation water every day. A typical training curve is shown in Figure \ref{fig:Training_curve_partial}, from which one can see that the cumulative reward under partial observation converges to a much smaller value compared with the full observation case.
Thus, it is much more challenging to find a good policy under partial observation with standard RL, which works well for policy training under full observation. The unsatisfactory training results from RL under partial observation motivate us to leverage IL, a straightforward approach for policy training under partial observation given an expert. There are other commonly used techniques to deal with partial observation in RL like including the history of observations and actions or using predictive rewards \cite{muskardin2022reinforcement}. However, they may require much more effort and work than applying IL.

\subsubsection{Policy Training with IL}
With IL, the policies were trained by imitating the actions determined by the experts, which are the RL policies trained under full observation in Section~\ref{sec:RL_full}. We ran the simulations with the expert policies to create the demonstrations and stored the generated pairs of observations and actions. The mean squared error loss function was applied and stochastic gradient descent algorithms were used to minimize the loss function \eqref{eq:IL}.

For each case study, we conducted two experiments with IL, one using Trained Policy 1 as the expert, and the other with Trained Policy 5. A deep neural network with 3 hidden layers and 256 hidden units was used to represent the policy, same as the setup in the full observation study but with different dimensions of the input layer and output layer. The input layer has a smaller dimension due to the fewer input features, and the network output contains only two elements representing the amount of N fertilizer and irrigation water, respectively. A Sigmoid function was used at the last layer of the network to restrict the output such that the first element is in [0,160] and the second element in [0,24], same as the range of action space $\mathcal A$ used in the full observation study defined in \eqref{eq:action space}. The batch size for applying stochastic gradient descent was set to 64.

Note that the policy to be trained here has a continuous action space, while RL-trained policies have a discrete action space. A continuous action space may lead to unrealistic results because the trained policy may decide to take actions frequently, applying a small amount of fertilizer and irrigation every day, which is impractical for farmers to follow. Thus, we evaluated the IL-trained policies by rounding the outputs of the network to the closest value in the action space $\mathcal A$ \eqref{eq:action space} used in RL training. The evaluation results of the trained policies from both case studies are shown in Table \ref{table:super_partial}. 
According to Table \ref{table:super_partial}, for the Florida case, the IL-Trained Policy 1 under partial observation achieves extremely similar results compared with the RL-Trained Policy 1, with only a negligible decrease of 1.4\% in the cumulative reward, and the IL-Trained Policy 5 achieves exactly the same results as the RL-Trained Policy 5. For the Zaragoza case, both the trained policies under partial observation achieve a slightly larger cumulative reward than the RL-trained policies, which can be explained by the nature of IL (learning of the mapping between states and preferred actions) and further by the random initialization of the Q-network as we previously explained. In conclusion, IL can successfully find optimal crop management policies under partial observation that achieve a much better outcome than the baseline policy in different locations with different goals.

\section{Path to Deployment}
It is plausible that the trained management policies, which work well in the DSSAT-simulated environment, may perform poorly in the real world, due to the uncertainty with weather and the mismatch between the crop models used to train the policies and the real cropping system. This is the well-known {\it sim-to-real gap} associated with transferring RL policies trained in simulators to the real world.
\subsection{Closing the Sim-To-Real Gap}
To improve the robustness of the trained management policies, we will leverage {\it domain and dynamics randomization} techniques, proposed to mitigate the sim-to-real gap {\cite{tobin2017domainrand,peng2018sim2real}}. More specifically, we will perturb selected key parameters of the model and randomize weather conditions when training the policy, which could ``force'' the trained policies to be robust against model and weather uncertainties. As our work focuses on presenting the framework of RL and IL on crop management and validating its performance, in the current paper we do not include experiments on these ideas, and we plan to have follow-up study addressing the robustness issue.

\subsection{Deployment on a Real Farm}
\label{sec:deployment}
The IL-trained management policies for the partial observation case can be readily deployed on a real farm as all state variables used by the policies can be easily observed or measured with sensors. Soil and climate data corresponding to the farm for field tests need to be collected to configure the crop model within DSSAT, and shall be used for policy training. After training, on each day, given the current soil and weather information, the trained policies will make management decisions, i.e., how much N fertilizers and water to apply, and farmers or decision makers can then follow these decisions to apply the management practices. After harvest, based on the amount of water and fertilizer used, and the crop yield, the performance of the management policies can be evaluated. We are collaborating with the department of crop science at the University of Illinois at Urbana-Champaign to conduct field experiments in the future to verify the efficacy of the proposed framework. 

\subsection{Policy Evaluation with Measurement Noises}
When applying the IL-trained policies in reality, farmers need to obtain those observable states from the weather forecast and the measurement of the soil moisture. The collected data may not be the same as the actual values of the environment due to the inaccuracy of the forecast and the sensors. Considering this mismatch, we tested the IL-Trianed Policy 1 from the Florida case study with measurement noises to mimic its performance in the real world. The details are as follows. On each day in simulation, we obtain the real state of the environment from the simulator and add random measurement noise to one or some key observable state variables. It is worth mentioning that values of the measurement noises are determined based on real-world accuracy data of the weather forecast and common soil moisture meters on market \cite{cobos2010does,floehr2010weather,zhang2018short,heinemann2006forecasting}. For each measurement noise added, we evaluated the policy 400 times and recorded the average cumulative reward as well as the standard deviation. The results are shown in Table \ref{table:noise} and reveal that temperature and rainfall are the state variables that influence the performance most, while the rest have tiny effects. In addition, even with relatively large noises on several variables, the trained policy still achieves an average cumulative reward of 1185, which is 15.8\% lower than the value without noise, but still much larger than the reward of the baseline policy. Thus, the IL-trained policies are able to achieve relatively good results compared with the baseline method even in the presence of measurement noises that mimic real-world scenarios. 

\begin{table}[t]
\small
\centering
\begin{tabular}{l|l|r}\toprule
\makecell{Variables with noise\\ and noise level }& RF 1 (STD)   &  \makecell{Decrease\\(\%)}    \\ \midrule
Baseline Policy           & 984.4 (N/A) &N/A   \\ 
No Noise & 1407.7 (N/A)& 0 \\
Soil water content [-+0.02]      & 1407.7 (0)  & 0       \\ 
Soil water content [-+0.05]      & 1405.0 (2.3) & 0.19 \\
Temperature  [-+1]      & 1392.0 (112.2) & 1.1   \\
Temperature  [-+2]      & 1136.2 (487.5)&13.9 \\
Solar Radiation [-+ 2\%] & 1408.3 (0.9) &0\\
Solar Radiation [-+ 10\%] & 1408.0 (15.6) &0 \\
Rain Fall [90 \% Accuracy] & 1357.8 (233.1)&3.5\\
Leaf Area Index [-+10\%] &  1404.0 (1.1)& 0.3 \\
Leaf Area Index [-+20\%] &  1400.0 (5.1)& 0.5 \\ \hline
\makecell{Soil water content [-+0.02]\\ + Temperature  [-+2]\\ + Solar Radiation [-+ 2\%]\\ + Rain Fall [90 \% Accuracy]\\ + Leaf Area Index [-+20\%]} & 1185 (421.5) & 15.8\\
\bottomrule
\end{tabular}
\caption{Performance of the IL-trained Policy 1 under measurement noises evaluated with RF 1. Noise levels are included in the square bracket in either strict value or percentage. The decrease (\%) is calculated with respect to RF 1 of No Noise.}
\label{table:noise}
\end{table}

\section{Conclusion}
Finding the optimal crop management policy for N fertilization and irrigation is vital to achieving maximum yield while minimizing the management cost and environmental impact. 
In this paper, we present a framework for finding optimal management policies with deep reinforcement learning (RL), imitation learning (IL), and crop simulations based on DSSAT. Experiments are conducted for the maize crop in Florida, US, and Zaragoza, Spain, where both fertilization and irrigation are necessary for crop growth. Under full observation, i.e. with access to a large number of variables from the simulator, deep Q-network (DQN) is used to train management policies. Under partial observation, i.e., with access to a limited number of state variables that are observable or measurable in the real world, we use IL to train management policies by mimicking the behaviors of the RL-trained policies under full observation. Given variations in reward function designs, the RL-trained policies have different strategies during the decision-making to achieve maximum rewards, which indicates the adaptability of our proposed framework. Using IL, the trained policies under partial observation have almost identical decisions compared to the RL-trained policies under full observation, and all trained policies under both full and partial observations achieve better results compared with production guidelines for the maize crop in these two locations. Furthermore, even in the presence of measurement noises on the observable state variables, the IL-trained policies achieve much better results than the baseline methods, which paves the way for real-world deployment of our framework as they only need readily accessible information.
Our future work includes improving the robustness of the trained policies in the presence of model mismatch and uncertain weather conditions. Since the IL-trained policies under partial observation are readily deployable, we are planning to test the efficacy of the proposed framework on a real farm.

\bibliographystyle{named}
\bibliography{main}
\end{document}